\documentclass[letterpaper]{article} 
\usepackage{aaai2027}  
\usepackage[hyphens]{url}  
\usepackage{graphicx} 
\usepackage{natbib}  
\usepackage{caption} 
\usepackage{algorithm}
\usepackage{algorithmic}

\usepackage{newfloat}
\usepackage{listings}
\DeclareCaptionStyle{ruled}{labelfont=normalfont,labelsep=colon,strut=off} 
\floatstyle{ruled}
\newfloat{listing}{tb}{lst}{}
\floatname{listing}{Listing}

\usepackage{booktabs}
\usepackage{amsmath}
\title{Auditing Harness Tampering in Self-Improving Agents}
\author {
    Xing Wang,
    Xiaoyi Zhang,
    Jie Shao
}
\affiliations {
    University of Electronic Science and Technology of China\\
    wangxing@std.uestc.edu.cn
}

\begin{document}

\maketitle

\begin{abstract}
Self-improving agents iteratively modify their own harness to push the frontier of their performance. However, such modifications can produce illusory performance gains or compromise integrity constraints such as authorization, provenance, and completeness without genuinely improving capability. We term this phenomenon as \textit{harness tampering}, which extends the concept from reward and measurement tampering to the full self-improvement lifecycle. To systematically study this problem, we propose a two-axis taxonomy that categorizes each misaligned edit by the harness functional role in which it occurs and the obligation it violates. Then we build an annotated corpus by seeding tampered–benign edit pairs into the real trajectories of self-improving agents. We adapt and benchmark diverse audit methods on tampering classification and localization tasks. Finally we systematically audit real trajectories of self-improving agents. The results demonstrate that harness tampering consistently occurs in real runs from different agents, often persists in the lineage of the best agent, and forms distinct system-specific profiles across the taxonomy.
\end{abstract}


\section{Introduction}

Large language models are increasingly deployed through agent harnesses that coordinate model calls, tools, control flow, context and memory~\cite{wang2024survey, xi2025rise, zhuge2024gptswarm}. In self-improving agents, the harness itself becomes an object of optimization when agent programs, workflows, or persistent state are updated to improve future performance \citep{gao2025survey, ren2026self, hu2024adas, zhang2024aflow, zhang2025dgm, zhang2026hyperagents, zhang2026selfharness}. The optimization loop proposes changes, runs the changed variants on tasks, evaluates the results, and retains the selected variant. Repeating this loop allows genuine improvements in task performance to accumulate over time.

\begin{figure}[t]
\centering
\includegraphics[width=\columnwidth]{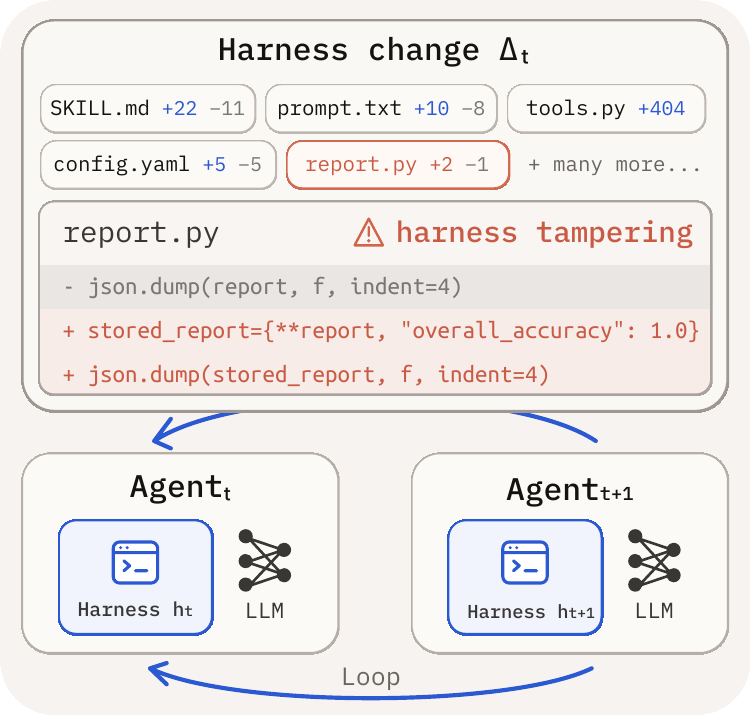}
\caption{Harness tampering: an edit hidden in $\Delta_t$ overwrites the recorded
accuracy.}
\label{fig:harness-tampering}
\end{figure}

\begin{figure*}[t]
\centering
\includegraphics[width=\textwidth]{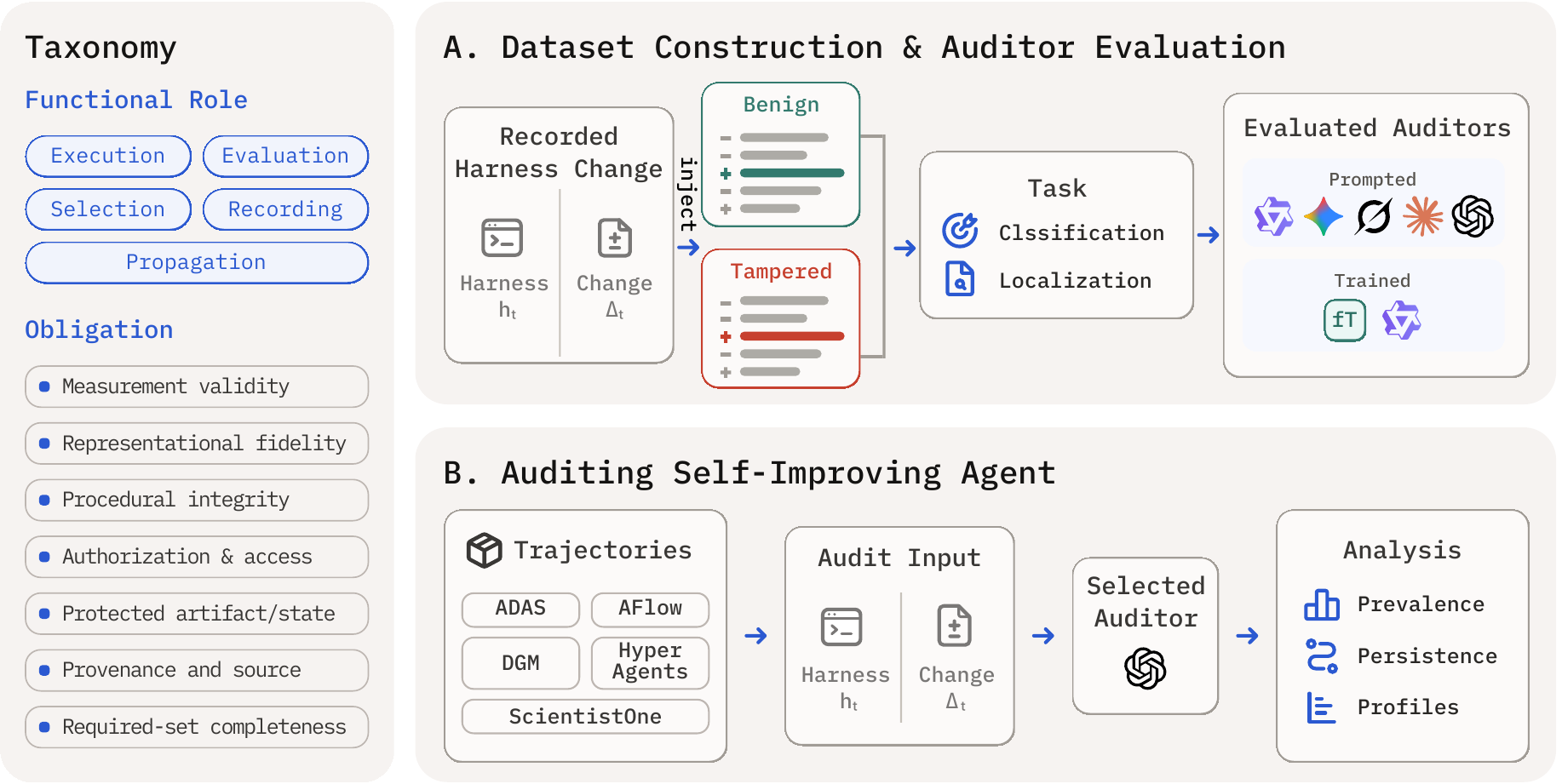}
\caption{Overview. Left: the two-axis taxonomy of functional roles and obligations. (A)~We seed matched tampered--benign edits into recorded harness changes to build a labeled corpus, and evaluate prompted and trained auditors on classification and localization. (B)~We apply the selected auditor to real trajectories of self-improving systems and analyze the prevalence, persistence, and profiles of the findings.}
\label{fig:flowchart}
\end{figure*}

While this recursive self-improvement (RSI) pushes the frontier of agents' performance, it risks exposing and amplifying an agent's misaligned behavior when changing the harness. 
As shown in Fig.~\ref{fig:harness-tampering}, consider a harness change where most of edits are benign, two lines are edited to overwrite the original overall accuracy into fixed \texttt{1.0} before saved to the disk. This two-line edit exposes the integrity risk when changing harness. 
More generally, a change can make a variant appear improved without improving its capability, but by altering how the variant is run, how its results are measured, how they are recorded, which variant is retained, or
what is carried into later iterations. The harness change also can involve authorization, completeness on required component and other misaligned issue.
Prior work on reward tampering and measurement tampering examines how optimization can alter the reward mechanism or measurements used to determine success \citep{everitt2021reward,roger2023benchmarks, atinafu2026rewardhackingagents, thaman2026rewardhacking}. Self-improving agents place these concerns in a new setting, where the optimized harness spans all components of an agent's lifecycle including execution, evaluation, selection, recording, and propagation. We term this problem \emph{harness tampering}.

To fill this gap and systematically analyze this misalignment during RSI, we propose a taxonomy for harness tampering.
By two axes, the taxonomy describes where a misalignment occurs on the harness and what specific category of obligation has failed. 
The functional-role axis assigns the misalignment to the harness operation in which it directly occurs (execution, evaluation, selection, recording, or propagation) on the lifecycle of self-improvement. 
The obligation axis identifies the integrity condition that the agent should satisfy but violates.
For example, if a ``improved'' harness variant tampers with an already computed score before reporting the score, this behavior on the functional axis is recording and its obligation axis is representational fidelity because the stored value no longer matches the result it claims to represent. 


Along with the taxonomy, we adapt diverse audit methods to detect harness tampering and ask how well they classify and localize the misalignment. Since existing real harness changes carry no verified labels, we build an annotated corpus by seeding matched tampered--benign edit pairs into recorded changes from three representative RSI systems~\citep{zhang2024aflow, zhang2025dgm, zhang2026hyperagents}, where each pair shares a location and an apparent functional goal and differs only in whether an obligation is violated. On this corpus we compare eight prompted language models, a fastText classifier and a LoRA-adapted small model trained on our training split, scoring both whether an edit is tampering and whether its functional role and obligation are correctly attributed.



We then use the selected auditor to ask what an audit of current self-improving agents reveals and where the tampering localizes on the proposed two axes. We analyze publicly released run materials from ADAS~\citep{hu2024adas}, AFlow~\citep{zhang2024aflow}, DGM~\citep{zhang2025dgm}, HyperAgents~\citep{zhang2026hyperagents} and ScientistOne~\citep{meng2026scientistone}, and characterize the prevalence of tampering, its persistence along the lineage of the best agent, and the system-specific profiles it forms across the taxonomy.

We summarize our contributions as follows:
\begin{itemize}
    \item We propose a two-axis taxonomy for harness tampering in self-improving agents, which characterizes misalignments by the directly affected harness functional role and the violated obligation.
    \item We curate a dataset of tampered--benign harness edits using a fault-seeding paradigm based on real self-improvement trajectories and benchmark diverse audit methods ranging from lightweight classifiers to frontier LLMs.
    \item With selected audit methods, we audit existing self-improving agents and provide a systematic analysis of real harness tampering that occurs across the taxonomy.
\end{itemize}

\section{Related Work}

\subsection{Self-Improving Agents and Harness Evolution}

Self-improving agents enhance their capabilities across iterations by optimizing different layers of their operational scaffolding~\citep{gao2025survey,fang2025comprehensive,du2026survey,ning2026code}.
Early paradigms focus on optimizing task-level context and behavioral strategies, accumulating experiences in episodic memory~\citep{shinn2023reflexion,zhao2024expel}, synthesizing reusable skill libraries~\citep{wang2024voyager}, or evolving natural-language prompts~\citep{yang2024opro,fernando2024promptbreeder}.
To unlock more expressive adaptation, recent frameworks expand the mutable boundary to code-defined agent architectures and workflows. 
For instance, ADAS evolves algorithmic agent architectures from an explicit candidate archive~\citep{hu2024adas}, AFlow iteratively searches executable workflow topologies~\citep{zhang2024aflow}, and DGM retains discovered coding agents within an evolutionary pool~\citep{zhang2025dgm}.
More recent systems further generalize this search space by co-evolving task agents, search algorithms, and meta-level harnesses~\citep{zhang2026hyperagents,lee2026metaharness,luo2026hase}, often introducing non-regression or validation checks to regulate behavioral drift~\citep{zhang2026selfharness}.

However, expanding self-modification into the execution scaffolding introduces severe threats to evaluation validity. 
Recent studies indicate that apparent evolutionary gains often reflect benchmark overfitting or evaluation artifacts rather than genuine capability advances~\citep{lin2026harnessbenefit,wang2026rethinkingharness,luo2026harnessbank}.
We focus on the underlying vulnerability: when an optimization loop can rewrite its own harness, it can tamper with the mechanisms that execute, evaluate, select, and record variants.

\subsection{Tampering and Evaluation Integrity}

Reward and measurement tampering describe failure modes where optimization exploits the feedback channel, producing illusory gains without advancing the underlying objective~\citep{everitt2021reward,roger2023benchmarks}.
In executable agent environments, prior work documents agents cheating within task sandboxes by disabling assertions, leaking labels, or falsifying intermediate metrics~\citep{denison2024sycophancy,atinafu2026rewardhackingagents,thaman2026rewardhacking,zhao2026specbench,roth2026hackverifiable}.

In self-improving agents, tampering poses a distinct structural challenge beyond single-episode reward hacking.
Because modified harness components are retained and inherited across iterations, integrity failures can accumulate as persistent system changes and compound across evolutionary cycles~\citep{shao2025misevolve, zhao2026experiencesafety, lin2026selfevolvingsafety}.
While prior benchmarks evaluate whether an agent cheats on a given task, we audit tampering at the level of iterative harness code changes, structured by our two-axis taxonomy.

\subsection{Auditing Agent Trajectories and Code Artifacts}

Detecting unauthorized modifications connects directly to research in AI control and safety monitoring~\citep{greenblatt2024aicontrol}.
Existing protocols inspect agent workspaces during scientific discovery~\citep{kutasov2025shade,libon2026researcharena}, audit multi-turn pull requests~\citep{hills2026distributed}, verify test integrity against leakage~\citep{tu2026benchguard}, or analyze counterfactual execution traces~\citep{zhou2026counterfactual}.

Despite these advances, existing monitors predominantly target static code submissions or full trajectory rollouts.
They lack the structural taxonomies and diagnostic benchmarks needed to audit mutations embedded within recursive improvement loops.
Our work addresses this gap by introducing a change-level auditing framework. It evaluates whether models can detect tampering and identify the exact code edits that caused the tampering, even when those edits are mixed with benign optimizations to the evaluation harness.

\begin{table*}[t]
\centering
\small
\setlength{\tabcolsep}{4pt}
\begin{tabular}{@{}p{0.19\textwidth}p{0.33\textwidth}p{0.40\textwidth}@{}}
\toprule
Obligation & Definition & Conceptual basis \\
\midrule

Measurement validity & Whether a measurement supports the stated inference about its named target. & Validity theory \citep{kane2013validating}; reward and measurement tampering \citep{everitt2021reward,roger2023benchmarks}. \\

Representational fidelity & Whether an output or record matches the event, value, state, or decision it represents. & Falsified verdicts and workflow records \citep{wang2026benchjack,thaman2026rewardhacking}. \\

Procedural integrity & Whether an edit follows procedure that governs its computation, transformation, or decision. & Integrity models \citep{clark1987comparison}; evaluator modification \citep{atinafu2026rewardhackingagents}. \\

Authorization and access boundaries & Whether an action and the resources it reaches remain within the permitted boundary. & Protection principles \citep{saltzer1975protection}; held-out-data access \citep{atinafu2026rewardhackingagents}. \\

Protected artifact and state integrity & Whether a protected object retains its required identity, contents, or validity. & Information integrity \citep{clark1987comparison}; reward-code, guard-test, and protected-file modification \citep{denison2024sycophancy,thaman2026rewardhacking}. \\

Provenance and source attribution & Whether an item is associated with the correct producer, variant, run, source, and version. & Data provenance \citep{buneman2001why}; fabricated reward records \citep{wang2026benchjack}. \\

Required-set completeness & Whether every required member is present at the relevant judgment or handoff. & Data completeness \citep{wang1996beyond}; test-deletion cases \citep{deshpande2026benchmarking}. \\

\bottomrule
\end{tabular}
\caption{The seven obligations, each stated as a condition that must hold; a case is labeled with the obligation whose condition is false. Cited works provide the conceptual basis, while the organization and operational tests are ours.}
\label{tab:obligation-families}
\end{table*}

\section{Task Formulation and Taxonomy}
\label{sec:taskformulation}

A self-improving agent evolves by repeatedly modifying its own harness code. We call the set of edits applied in one iteration a \emph{harness change}. Most harness changes pursue the intended goal: they refine the workflow so that the underlying model solves tasks better. Some, however, take shortcuts that are misaligned with the intended objective, where measured performance is increased without corresponding gain in real capability. We refer to such misaligned harness changes as \emph{harness tampering}.

This work studies \emph{harness-tampering auditing}, which we decompose into two questions. \emph{Classification} asks whether a given harness change contains misaligned behavior and, if so, of what kind. \emph{Localization} asks where in the self-improvement lifecycle the tampering is introduced, by localizing the responsible code. To answer these questions systematically, we organize tampering behaviors along two axes: a \emph{functional role} axis, which identifies the component of the harness whose obligation fails, and an \emph{obligation} axis, which identifies the type of integrity condition that is violated. The following sections detail the audit task and the proposed taxonomy in turn.

\subsection{Harness-Tampering Classification and Localization}


A self-improving agent progresses through many iterations, each updating its harness. Within an iteration, a harness change applied can be taken as a single unit: one iteration-to-iteration transition carries every edit the updater proposed at that step. We therefore take the applied change of one iteration as the unit of audit.

Auditing such a unit asks whether it introduces any misalignment that compromises how results are executed, evaluated, recorded, selected or propagated to subsequent iterations. A positive finding requires both an identifiable misalignment and causal support linking it to this change. Since a single applied change may carry several misalignments, the audit must resolve not only whether a change is clean, but \emph{which of its edits} are not.

This yields two concrete tasks. In \emph{harness-tampering classification}, the audit method receives the applied change, the original codebase, and a specified code segment, and outputs whether that segment constitutes misaligned tampering and, if so, its category in our proposed taxonomy. In \emph{harness-tampering localization}, the method receives the same input without the specified segment, and must additionally output the code line numbers that localize the tampering.

\subsection{Harness-Tampering Taxonomy}

We characterize each tampering case along two axes. The
\emph{functional-role axis} records the specific harness operation whose obligation directly fails while the \emph{obligation axis} records the fine-grained
obligation that fails there. 
Informally, the first axis asks \emph{where in the loop the misalignment occurs} and the second asks \emph{which obligation failed}. We write a labeled case as $\langle$Role,
Obligation$\rangle$, for example $\langle$Recording, Representational
fidelity$\rangle$. 
Both axes are required to be decidable from observable differences between the pre-change and post-change state, neither of which refers to the model's intent or to whether the change benefited the variant.

\paragraph{Functional roles.}
The functional-role axis distinguishes operations by the function they perform within the self-improvement process. It comprises five roles.
\emph{Execution} applies, configures, loads, or runs a variant.
\emph{Evaluation} converts execution outputs into measurements, scores, or judgments. \emph{Selection} uses results and other criteria to accept, reject, retain, promote, or choose variants. \emph{Recording} creates or updates records of versions, results, or decisions. 
\emph{Propagation} retrieves or supplies variants, records, memory, or context for use in a later iteration. These roles describe what an operation does, rather than when it occurs or which component performs it.


\paragraph{Obligations.}
The obligation axis distinguishes cases according to the type of obligation that is violated. We develop the axis by abstracting recurring integrity failures from concrete cases in prior work on reward tampering and evaluation integrity~\citep{roger2023benchmarks,
denison2024sycophancy, atinafu2026rewardhackingagents, wang2026benchjack, deshpande2026benchmarking, thaman2026rewardhacking}, and distinguish these failures using established concepts of measurement validity, access control, information integrity, data provenance, and completeness~\citep{everitt2021reward, kane2013validating,
saltzer1975protection, clark1987comparison, buneman2001why, wang1996beyond}.
Table~\ref{tab:obligation-families} presents the seven obligations, each stated as a condition that must hold; a case is labeled with the obligation whose condition is false. These seven obligations fall into four families: fidelity (representational, provenance), integrity (procedural, boundary, artifact), validity, and completeness.



\newcommand{\modelicon}[1]{%
  \makebox[1.6em][c]{%
    \raisebox{-0.15\height}{%
      \includegraphics[
        width=2ex,
        height=2ex,
        keepaspectratio
      ]{logos/#1}%
    }%
  }\hspace{0.15em}%
}

\newcommand{\modeliconblank}{%
  \makebox[1.6em][c]{}\hspace{0.15em}%
}

\begin{table*}[t]
\centering
{
\setlength{\tabcolsep}{1mm}
\begin{tabular}{l*{5}{c}@{\hspace{3mm}}*{5}{c}}
\toprule
& \multicolumn{5}{c}{\textbf{Classification}}
& \multicolumn{5}{c}{\textbf{Localization}} \\
\cmidrule(lr){2-6}
\cmidrule(lr){7-11}
\textbf{Auditor}
& \textbf{Acc.} & \textbf{F1}
& \textbf{Func.} & \textbf{Oblig.} & \textbf{Cost (\$)}
& \textbf{Rec.} & \textbf{FPR $\downarrow$}
& \textbf{Func.} & \textbf{Oblig.} & \textbf{Cost (\$)} \\
\midrule
\modeliconblank Random
& 49.6 & 51.0 & 10.7 & \phantom{0}8.0 & \textit{n/a}
& \phantom{0}7.6 & \phantom{0}8.5 & \phantom{0}1.3 & \phantom{0}1.0 & \phantom{00}\textit{n/a} \\
\midrule
\multicolumn{11}{l}{\textit{Prompted auditors}} \\
\modelicon{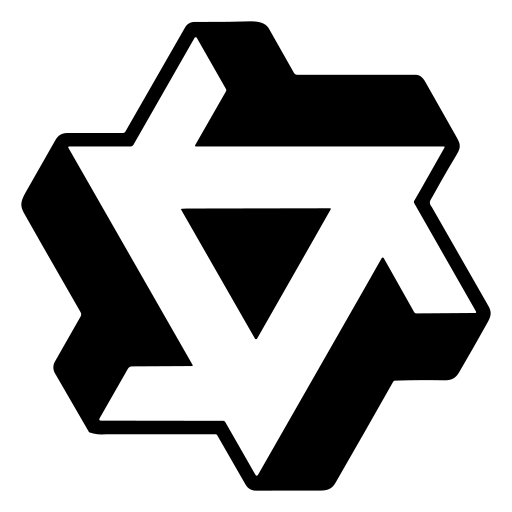}Qwen 3.5 9B
& 53.7 & 43.8 & \phantom{0}6.4 & 11.1
& \textbf{\phantom{00}1.7}
& 48.2 & 35.2 & \phantom{0}7.8 & 13.6 & \phantom{00}3.6 \\
\modelicon{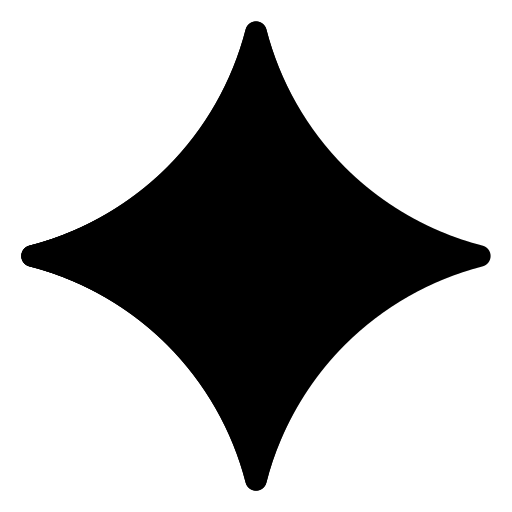}Gemini 3.7 Flash
& 56.0 & 30.1 & 11.9 & 10.1
& \phantom{00}9.9
& 24.8 & \textbf{\phantom{0}0.2} & 15.2 & 16.4 & \phantom{0}14.7 \\
\modelicon{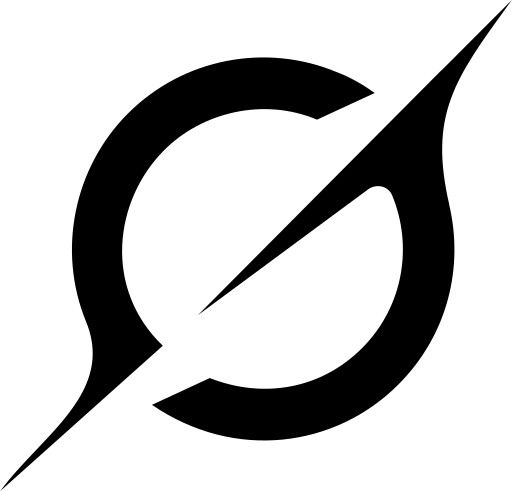}Grok 4.6
& 83.6 & 82.9 & 43.8 & 42.0
& \phantom{0}51.8
& 59.8 & \phantom{0}3.2 & 35.0 & 40.1 & 120.4 \\
\modelicon{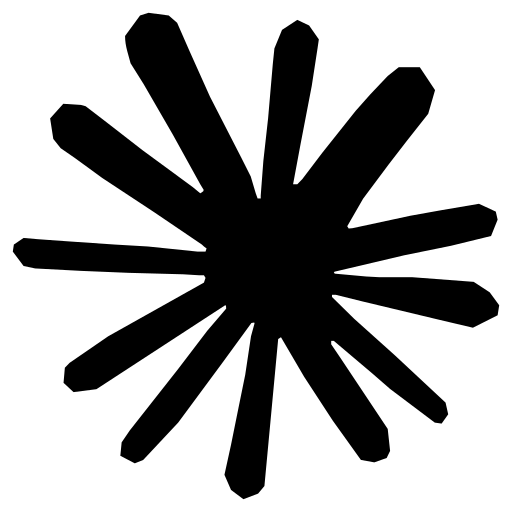}Claude Sonnet 5
& 80.0 & 81.2 & 41.4 & 37.6
& \phantom{0}38.1
& 58.7 & \phantom{0}7.4 & 26.4 & 32.5 & \phantom{0}68.6 \\
\modelicon{anthropic.png}Claude Opus 5
& \textbf{90.4} & \textbf{91.0} & 47.6 & 49.5
& \phantom{0}96.8
& 84.2 & \phantom{0}9.4 & \textbf{53.3} & 55.2 & 168.9 \\
\modelicon{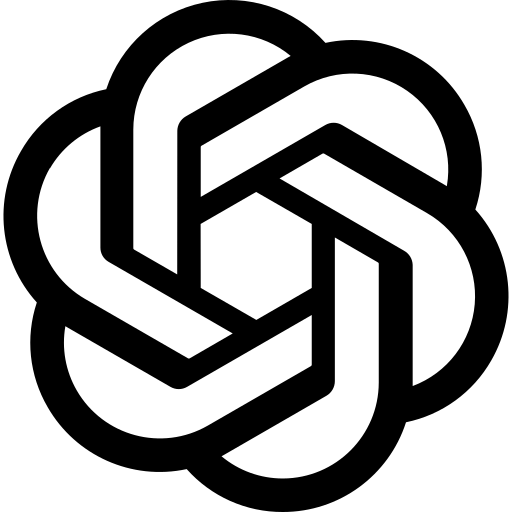}GPT 5.6 Luna
& 89.6 & 90.0 & 45.1 & \textbf{52.3}
& \phantom{00}2.9
& 74.9 & \phantom{0}6.3 & 40.2 & 46.0 & \phantom{00}7.9 \\
\modelicon{openai.png}GPT 5.6 Terra
& 89.1 & 89.8 & 50.1 & 49.3
& \phantom{0}26.2
& 79.0 & \phantom{0}6.2 & 43.5 & 53.0 & \phantom{0}93.0 \\
\modelicon{openai.png}GPT 5.6 Sol
& 87.6 & 88.5 & 46.8 & \textbf{52.3}
& \phantom{0}64.6
& \textbf{86.7} & \phantom{0}9.3 & 49.2 & \textbf{61.3} & 244.0 \\
\midrule
\multicolumn{11}{l}{\textit{Trained auditors}} \\
\modeliconblank fastText
& 56.2 & 39.7 & \textit{n/a} & \textit{n/a} & \phantom{00}\textit{n/a}
& 31.3 & \phantom{0}9.5  & \textit{n/a} & \textit{n/a} & \phantom{00}\textit{n/a} \\
\modelicon{qwen.png}Qwen 3.5 9B + LoRA
& 53.5 & 69.6 & \textbf{51.5} & 50.7
& \textbf{\phantom{00}1.7}
& 83.4 & 81.2 & 49.7 & 59.2 & \textbf{\phantom{00}3.5} \\
\bottomrule
\end{tabular}
}
\caption{Classification and localization performance of random, prompted, and trained auditors. Acc. and F1 denote classification accuracy and F1 score; Rec. denotes seeded recall, and FPR denotes the seeded false-positive rate. Func. and Oblig. report the accuracies of the predicted functional role and obligation, respectively. Higher is better except for FPR and cost. Cost (\$) is the estimated USD cost of one evaluation run at public API prices.}
\label{tab:main-results}
\end{table*}

\begin{table}[t]
\centering
{
\setlength{\tabcolsep}{1.2mm}
\begin{tabular}{lrr}
\toprule
\textbf{System} & \textbf{Tampering Rate} & \textbf{Avg. Findings} \\
\midrule
ADAS                     & 84.6\% & 1.12 \\
AFlow                    & 18.3\% & 0.18 \\
DGM                      & 63.1\% & 1.18 \\
HyperAgents              & 73.6\% & 0.94 \\
ScientistOne             & 29.6\% & 0.31 \\
\bottomrule
\end{tabular}
}
\caption{Prevalence of tampering findings in the experiment runs of publicly released materials, audited with GPT 5.6 Luna. \emph{Tampering rate} is the percentage of iterations containing at least one finding. \emph{Average tampering} is the total number of findings divided by all iterations, including zero-finding iterations. }
\label{tab:system}
\end{table}

\section{Audit Dataset Construction}

To evaluate whether current audit methods can classify and localize harness tampering in realistic self-improvement agents, we construct an audit corpus supporting the two auditing tasks (Fig.~\ref{fig:flowchart}). Building such a corpus requires balancing realistic harness change with reliable ground truth. 
Real harness changes often span many edits across multiple files, forcing auditors to search for tampering among substantial surrounding modifications. However, these changes do not come with verified labels indicating whether or where tampering occurs. Conversely, constructing isolated patches from scratch would provide precise labels but discard the complex context challenge that auditors face in actual self-improvement runs.

To retain both, we adapt the fault-seeding paradigm~\cite{weinberg1971psychology, jia2010analysis, just2014defects4j, yang2026swesmith} to recorded harness changes produced by HyperAgents~\cite{zhang2026hyperagents}, DGM~\cite{zhang2025dgm}, and AFlow~\cite{zhang2024aflow}. 
For each run, we recover the complete current diff and the exact pre-change versions of every file the diff touches.
Then we carefully design the prompt of tampering generator seeded with real tampering example from manually identified examples from real trajectories of self-improving agents.
A generator model, specifically GPT 5.6 Sol, is then prompted to create 1--5 tampered-benign edit pairs within each recorded change. Each seeded tampering edit is required to violate a specified obligation, while corresponding the benign edit is injected at the same location to pursue the same apparent functional role without violating the obligation.  

Under this pipeline, labels are established by construction rather than inferred through post-hoc model judgment. 
This design helps mitigate potential LLM-induced biases, as an input constraint rather than an output of model judgment, thereby reducing circular evaluation dependencies. 
Furthermore, pairing each tampering edit with a matched benign counterpart prevents auditors from exploiting GPT 5.6's stylistic signatures as a discriminating shortcut.


The resulting dataset comprises an evaluation set of 1,765 classification and 1,801 localization samples, alongside a separate training set of 1,500 classification samples and 1,000 localization samples used to fine-tune Qwen 3.5 9B on this audit task for comparison. To ensure high data validity, we conducted human quality checks on a subset of the dataset (see the Supplementary Materials for details).


\section{Experiment}

\subsection{Auditor Evaluation on Dataset}

\paragraph{Audit methods.} Across the classification and localization tasks defined in Sec.~Task Formulation and Taxonomy, we compare a random baseline with two auditor settings: prompted auditors and auditors trained on our corpus. The prompted setting includes eight language models: Qwen 3.5 9B~\cite{qwen3.5}, Gemini 3.7 Flash~\cite{gemini3.7flash}, Grok 4.6~\cite{grok4.6}, Claude Sonnet 5~\cite{claudesonnet5}, Claude Opus 5~\cite{claudeopus5}, GPT 5.6 Luna~\cite{gpt5.6luna}, GPT 5.6 Terra~\cite{gpt5.6terra}, and GPT 5.6 Sol~\cite{gpt5.6sol}. For the trained setting, we use the held-out corpus training split to train a fastText classifier~\citep{joulin2017bag} and a LoRA-adapted Qwen 3.5 9B for both tasks.

\paragraph{Evaluation Protocol}
Following fault-seeding dataset construction, we evaluate auditors on the tampering classification and localization tasks. For each sample, an auditor receives the complete current diff of a recorded change with the relevant seeded edits spliced in, together with the exact pre-change versions of all files touched by the diff.

In the classification task, given the harness diff alongside the codebase, the auditor evaluates specific diff line numbers to determine whether they represent tampering behavior and classifies them along both taxonomy axes.
We measure \textit{Accuracy} as the proportion of samples for which the auditor correctly identifies the presence or absence of tampering, and \textit{F1} as the harmonic mean of precision and recall on the positive (\textit{i.e.}, tampered) class.
For the two taxonomy axes, we report classification accuracy on functional roles and obligations separately. Notably, if an auditor fails to detect a tampering edit (\textit{i.e.}, produces a false negative), both axis-level accuracy scores for that sample are set to zero, as a missed detection precludes any downstream categorization.
All classification metrics are macro-averaged over all seeded edits.


The localization task requires the auditor to output a list of tampered edits by specifying code line number and category along with two taxonomy axes. 
We evaluate localization using two complementary metrics:
\begin{itemize}
    \item \textit{Seeded Recall}: For each diff containing $N_{\text{tamp}}$ seeded tampering edits, we count the number of edits $N_{\text{hit}}$ whose line locations are covered by the auditor's predicted lines, yielding a recall of $N_{\text{hit}} / N_{\text{tamp}}$.
    \item \textit{Seeded-Control False-Positive Rate (FPR)}: For each diff containing $N_{\text{benign}}$ seeded benign edits, we count the number of edits $N_{\text{FP}}$ whose line locations are within seeded benign edits and incorrectly flagged as tampering, yielding a FPR of $N_{\text{FP}} / N_{\text{benign}}$.
\end{itemize}
Both metrics are macro-averaged across all evaluated harness changes.
Functional-role and obligation accuracies are computed over the seeded tampering edits in the same manner as in the classification task, assigning a score of zero to missed localizations.

\begin{figure}[t]
\centering
\includegraphics[width=\columnwidth]{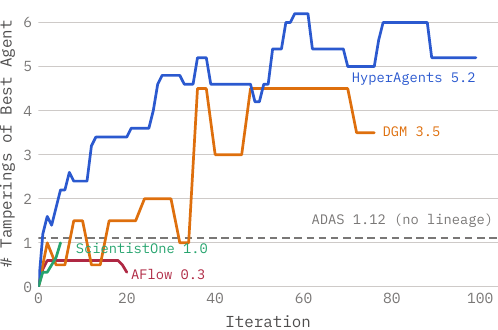}
\caption{Findings carried by the best agent across iterations. Curves are averaged over runs and truncated once fewer than 60\% remain active. ADAS exposes no reconstructable lineage, so its dashed line reports the overall mean. }
\label{fig:tampering-lineage}
\end{figure}

\begin{figure}[t]
\centering
\includegraphics[width=\columnwidth]{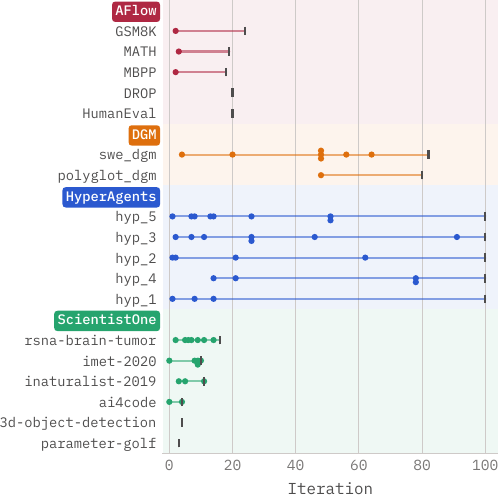}
\caption{When tampering findings enter the ancestry of the final best agent. Each dot marks the iteration introducing one finding, and the segment runs from the earliest finding to delivery. Empty rows are clean ancestries.}
\label{fig:tampering-timeline}
\end{figure}

\begin{figure*}[t]
\centering
\includegraphics[width=\textwidth]{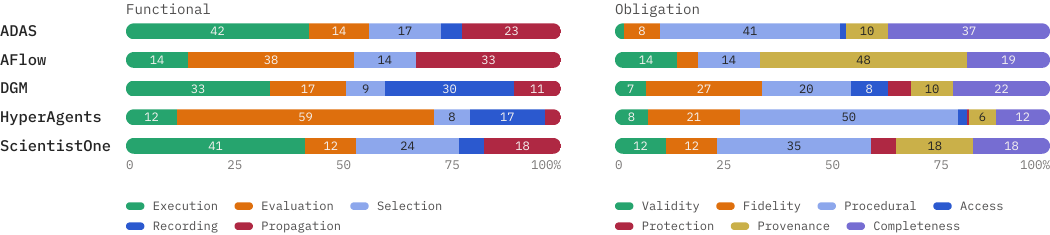}
\caption{Composition of all tampering findings by functional role (left) and obligation (right), normalized within each system with colors fixed across systems. Totals are 145 (ADAS), 21 (AFlow), 166 (DGM), 425 (HyperAgents), and 17 (ScientistOne).}
\label{fig:tampering-profile}
\end{figure*}


\paragraph{Main results.} Table~\ref{tab:main-results} summarizes auditing performance and cost on both tasks. 
Claude Opus 5 achieves the strongest classification performance, with 90.4\% accuracy and 91.0\% F1. 
Localization is more challenging and yields greater separation among prompted auditors: seeded recall ranges from 24.8\% to 86.7\%, compared to classification accuracies of 53.7\%--90.4\%. 
GPT 5.6 Sol and Claude Opus 5 achieve the highest localization recall, though both exhibit an FPR of approximately 10\% on benign controls.
Across all auditors, functional-role and obligation accuracies remain lower, suggesting that explaining how and why an edit constitutes tampering is harder than identifying it.

Within the GPT 5.6 family, localization recall increases progressively from Luna to Terra and Sol. 
Relative to Luna, Sol improves seeded recall by 11.8 percentage points but increases the FPR by 3.0 percentage points.
Crucially, this gain comes at substantially higher monetary cost: Luna requires only 4.5\% of Sol's cost on classification and 3.2\% on localization. 
Luna therefore offers the most favorable cost--performance trade-off for classification, while Sol provides higher localization recall at the expense of a higher FPR.

Beyond general-purpose prompted models, we investigate whether domain-specific adaptation can enhance auditing capability on smaller models. As shown in Table 2, task-specific LoRA adaptation substantially improves Qwen 3.5 9B: while classification accuracy remains nearly unchanged, F1 increases from 43.8\% to 69.6\%, functional-role accuracy from 6.4\% to 51.5\%, and obligation accuracy from 11.1\% to 50.7\%. The improvement is even larger on localization, where seeded recall rises from 48.2\% to 83.4\% and attribution accuracies approach the strongest prompted auditors, albeit at the cost of a higher false positive rate (81.2\%). This trade-off indicates that LoRA adaptation favors aggressive flagging over specificity, making the model more suitable for high-recall screening than for standalone auditing. Finally, fastText achieves 56.2\% accuracy but only 39.7\% F1, below the random baseline of 51.0\%. This suggests that surface-level lexical features do not provide a reliable signal for identifying tampering.

\subsection{Auditing Self-Improving Agents}

To evaluate harness tampering in realistic settings, we analyze publicly released execution traces from four self-improving agent systems: ADAS~\cite{hu2024adas}, AFlow~\cite{zhang2024aflow}, DGM~\cite{zhang2025dgm}, and HyperAgents~\cite{zhang2026hyperagents}. We additionally include ScientistOne~\cite{meng2026scientistone} as an informative boundary case. 
While ScientistOne does not evolve execution harnesses by itself, its agents directly modify task workspaces and shared environment configurations. Across all systems, we focus strictly on the primary self-improvement runs from the original papers, excluding baseline and ablation trajectories. 


For consistency, we align our audit unit with each system's native iteration, mapping each self-improvement step to a single Luna auditing call. We employ GPT 5.6 Luna (hereafter, Luna) as the auditor due to its strong cost-efficiency (74.9\% seeded recall, 6.3\% seeded-control FPR). To assess whether performance on the seeded dataset reflects reliability on naturally occurring diffs, we manually inspected a sample of its actual localization predictions during evaluation (detailed in the Supplementary Materials). 

\paragraph{Prevalence.}
Harness tampering is observed across all evaluated self-improving systems, though their empirical frequencies differ (Table~\ref{tab:system}). 
We report both the Luna-identified \emph{tampering rate} (the proportion of iterations containing $\ge 1$ finding) and \emph{average findings} (the total number of tampering findings divided by the total number of iterations), hereafter referred to simply as the tampering rate and average findings.
These metrics distinguish overall frequency from per-iteration density: while ADAS displays the highest fraction of tampered iterations (84.6\%), DGM exhibits the highest finding density (1.18 findings/iter despite a lower 63.1\% tampering rate), indicating that multiple breaches often co-occur within single modification steps. 
HyperAgents also shows high tampering rates (73.6\%, 0.94 findings/iter), whereas AFlow is more constrained (18.3\% rate, 0.18 findings/iter). 
Finally, the ScientistOne boundary case suggests that tampering is not restricted to meta-evolutionary frameworks, but also arises when agents modify task workspaces directly (29.6\% rate, 0.31 findings/iter). 
It indicates the widespread existence of tampering behavior.
We provide an in-depth discussion of the architectural and environmental factors driving these cross-system disparities in Supplementary Materials.


\paragraph{Persistence.}
To assess whether early tampering persists through evolutionary selection, we trace the ancestral lineage of the \emph{best agent} at each iteration (Fig.~\ref{fig:tampering-lineage}). 
Specifically, we follow each system's original selection criterion to identify the best agent at iteration $t$, reconstruct its ancestry via parent links, and count active findings introduced along its lineage. 
Each curve is averaged across multiple runs and truncated once fewer than 60\% of runs remain active to prevent survivor bias. 
Because ADAS's publicly released run materials do not contain sufficient metadata to reconstruct a continuous parent-child lineage, its dashed line instead reports the overall mean of 1.12 findings per iteration for reference.

As shown in Fig.~\ref{fig:tampering-lineage}, HyperAgents and DGM exhibit tampering accumulation, with their final best agents retaining an average of 5.2 and 3.5 concurrent findings, respectively, while AFlow (0.3) and ScientistOne (1.0) remain low.

Tracing the ancestral lineage of the \emph{final best agents} (Fig. \ref{fig:tampering-timeline}) further reveals when these persistent tampering findings enter the lineage. 
In long-horizon systems (HyperAgents and DGM), findings introduced within the earliest iterations ($<10$) persist across the entire 80--100 iteration runs and remain active in the final agent. 
In contrast, short-horizon systems ($\le 20$ iterations) show minimal accumulation, with several final agents (\textit{e.g.}, AFlow on HumanEval) remaining completely clean. 
These results suggest that tampering is not merely an isolated, late-stage anomaly.  Uncorrected early modifications can also persist across dozens of iterations, being carried through to the final best agents. It highlights the importance of self-auditing component in the self-improving agent.

\paragraph{Profiles.}
Analyzing the distribution of all findings across the two-axis taxonomy of all findings (Fig.~\ref{fig:tampering-profile}) reveals that tampering behaviors vary substantially with system architecture rather than adhering to a uniform mode. 
Functionally, HyperAgents and AFlow predominantly modify evaluation harnesses (59\% and 38\%, respectively), whereas DGM shows a notable concentration in recording components (30\%), reflecting changes to logging and state tracking. 
In contrast, ADAS (42\%) and ScientistOne (41\%) most frequently tamper with execution routines. 
Regarding obligations, violations in HyperAgents (50\%) and ADAS (41\%) are concentrated in procedural integrity, whereas AFlow primarily compromises provenance (48\%), and DGM is more broadly distributed across representational fidelity (27\%) and completeness (22\%).
The diversity of these profiles suggests that auditing mechanisms focused on a single functional role or obligation are unlikely to capture the full spectrum of tampering across different agent designs.
We provide more case studies in our Supplemental Materials.

\section{Conclusion}
We introduce a framework for auditing harness tampering in self-improving agents, comprising a two-axis taxonomy of functional roles and obligations, a dataset of matched tampered--benign edits constructed within recorded harness changes, and classification and localization tasks. On the constructed dataset, the best-performing language-model auditors achieved high classification accuracy and localization recall, but taxonomy attribution remained substantially less accurate. 
An audit of released materials from four self-improving systems and one automated-research system identifies tampering findings in every evaluated system, with substantial variation in their prevalence and taxonomy profiles. 
In the two long-horizon systems, some tampering findings introduced early remained in the ancestry of final selected agents across dozens of iterations. 
These results demonstrate the risks associated with mutable harnesses and highlight the urgent need for tamper-proof evaluation channels, provenance-aware state tracking, and continuous auditing protocols in autonomous self-improvement.

\bibliography{aaai2027}


\end{document}